\documentclass[runningheads]{llncs}
\pdfoutput=1
\usepackage[utf8]{inputenc}
\usepackage{todonotes}
\usepackage{dsfont}
\usepackage{enumitem,amssymb}
\usepackage{graphicx} 
\usepackage{url}

\title{FixMatchSeg: Fixing FixMatch for Semi-Supervised Semantic Segmentation}
\author{Pratima Upretee, Bishesh Khanal}
\institute{NepAl Applied Mathematics and Informatics Institute for research (NAAMII), Kathmandu, Nepal}
\begin{document}

\maketitle

\begin{abstract}
    
Supervised deep learning methods for semantic medical image segmentation are getting increasingly popular in the past few years.
However, in resource constrained settings, getting large number of annotated images is very difficult as it mostly requires experts, is expensive and time-consuming.
Semi-supervised segmentation can be an attractive solution where a very few labeled images are used along with a large number of unlabeled ones.
While the gap between supervised and semi-supervised methods have been dramatically reduced for classification problems in the past couple of years, there still remains a larger gap in segmentation methods.
In this work, we adapt a state-of-the-art semi-supervised classification method FixMatch  to semantic segmentation task, introducing FixMatchSeg.
FixMatchSeg is evaluated in four different publicly available datasets of different anatomy and different modality: cardiac ultrasound, chest X-ray, retinal fundus image, and skin images.
When there are few labels, we show that FixMatchSeg performs on par with strong supervised baselines.

\end{abstract}

\section{Introduction}
{\label{sec:Introduction}}

Almost all of recent studies in medical image segmentation focus on evaluating existing or developing new Deep Learning (DL) based models.
However, there are not many models that are actually deployed in real-world clinical settings.
With lack of experts, low-resource settings is where DL models could have a higher impact.
But in low-resource settings where annotation costs can be prohibitive, deploying DL models seems to be even further away as most of the successful DL models perform very poorly in low data regime.
Thus, developing methods that are accurate and robust with very few labels could help improve the accessibility and quality of healthcare services, especially in Low and Middle-Income Countries (LMICs).

Medical images usually have smaller number of annotated images compared to natural images datasets such as ImageNet.
Thus, transfer learning is quite common in medical image segmentation problems where a pre-trained network in datasets like ImageNet is fine-tuned in medical images.
Pretraining benefits only up to a certain level, and usually still requires reasonably large number of labeled medical images for fine-tuning.
Since the domain where pretraining is done is different from the target domain, without strong domain adaptation methods, it is difficult to increase the improvement brought about by transfer learning.
An attractive alternative is semi-supervised learning, where we have few labeled images and a large number of unlabeled images of the same domain.

Recently, a number of semi-supervised learning methods have been developed for classification tasks showing great promise and getting very close to the supervised learning performance \cite{sohn2020fixmatch,chen2020simple,grill2020bootstrap}.
These recent powerful methods primarily apply consistency regularization exploiting the smoothness or cluster assumption, i.e., same class images lie closely together in a manifold, and different classes are separated by decision boundaries that lie in low-density regions.
However, the semantic segmentation task has not seen similar success from semi-supervised learning yet.
In medical imaging, consistency regularization under transformation has been applied in some of the recent works \cite{Bortsova_2019,Li_2020,Perone_2018}.
But, most of them either use student and teacher models or ensemble models, increasing model complexity.
There is a lot of diversity within medical image segmentation task as there are several modalities with very different properties, anatomy regions of diverse size and shape, and different levels of texture vs shape bias requirement (irregular tumors or lesion vs heart or lungs having regular shape across samples).
It is not clear if the more complex architectures provide better performance across different types of medical imaging datasets or not compared to simpler approaches such as FixMatch \cite{sohn2020fixmatch}.
In resource constrained settings, simpler and smaller models requiring lower computational cost are more attractive.
Therefore, we adapt a simple but powerful end-to-end semi-supervised classification network for semantic segmentation of medical images and evaluate it across different modalities and anatomies.

We introduce FixMatchSeg\footnote{We introduced FixMatchSeg on July 2, 2021, the first version of this paper, as part of a submission to a conference}, a simple semi-supervised semantic segmentation method that exploits consistency regularization and pseudo-labeling \cite{Lee2013PseudoLabelT}.
It builds upon the recent progress in semi-supervised learning for classification tasks and adapts state-of-the-art FixMatch \cite{sohn2020fixmatch} to semantic segmentation tasks. 
We apply the proposed method on four different publicly available datasets: a) CAMUS \cite{Leclerc_2019} - set of echocardiographic images, b) REFUGE  \cite{ORLANDO2020101570}- dataset consisting of retinal fundus images, c) ISIC 2017 \cite{DBLP:journals/corr/abs-1710-05006}- dataset of dermoscopic images, and d) SCR \cite{VANGINNEKEN200619} (chest X-ray segmentation on JSRT database \cite{Shiraishi2000}) to check the dataset dependency of the proposed model. 

\paragraph{\textbf{Contributions:}}
\begin{itemize}
    \item Adapt a powerful state-of-the-art semi-supervised classification algorithm and extend it to medical image segmentation,
    
    \item Compare the performance of a semi-supervised model with a strong supervised learning baseline with both training from scratch and transfer learning scenario at different number of labeled images,
   
    \item Evaluate on four different publicly available datasets, showing that the proposed method provides results superior to supervised baseline when only few annotated labels are available. 
    
\end{itemize}
 
\begin{figure}[hbt!]
 \centering
    \includegraphics[width=1 \textwidth]{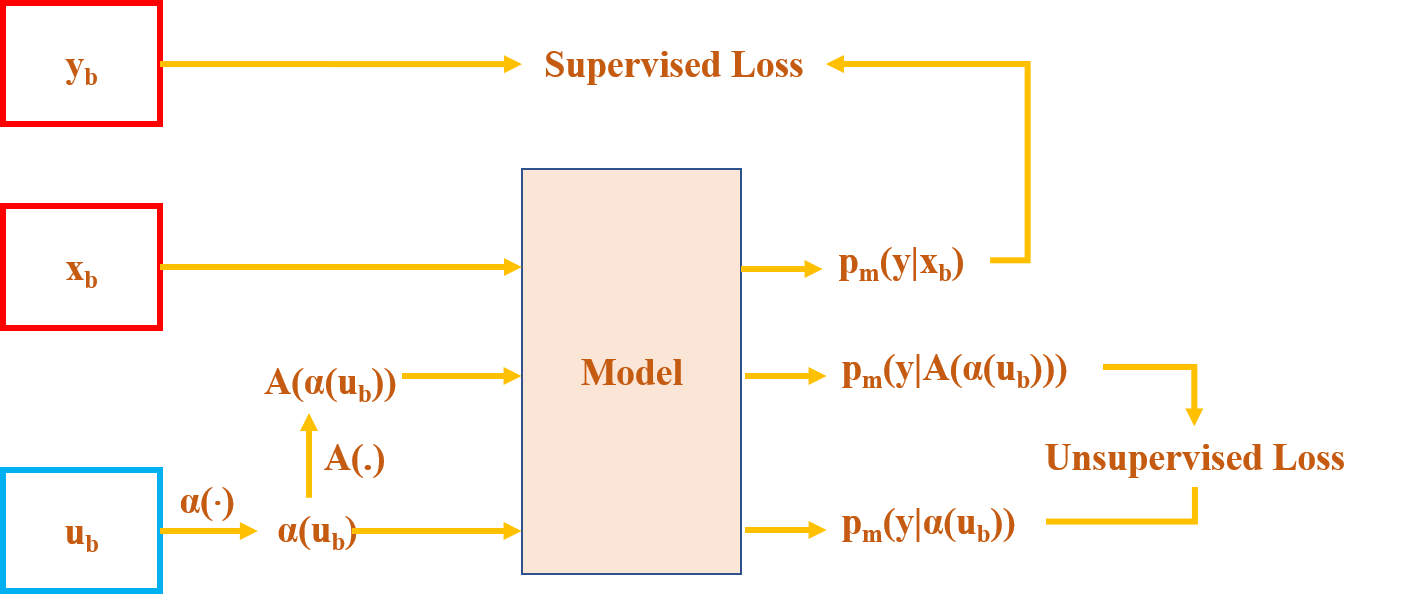}
    \caption{\textbf{Block diagram of proposed FixMatchSeg. }}
    \label{fig1}
\end{figure}

\section{Related Work}
{\label{sec:RelatedWork}}
Semi-supervised learning methods may be broadly classified into four categories: Consistency Regularization, Proxy-label and Entropy Minimization, Generative Models and Graph-Based Methods \cite{ouali2020overview}.
All these different methods have been explored for medical image segmentation such as consistency regularization in \cite{Baur_2017,Bortsova_2019,Cui_2019,Yu_2019,Li_2020},
proxy-label update for unlabeled images with alternate updates of
network parameters in \cite{Bai_2017}, generative models
in \cite{Sedai_2017,Zhang_2017,Nie_2018,Chartsias_2018} and graph-based methods in \cite{Ganaye_2018}.
These methods have been applied to a wide range of 2D and 3D medical images such as cardiac MRI \cite{Bai_2017,Chartsias_2018,Yu_2019}, brain lesion segmentation in MRI \cite{Baur_2017,Cui_2019}, chest X-ray \cite{Bortsova_2019}, skin lesion \cite{Li_2020}, retinal fundus images \cite{Sedai_2017}.

Several recent semi-supervised semantic segmentation methods for medical images propose consistency regularization with some sort of transformation consistency.
For example, \cite{Bortsova_2019} train Siamese network under transformation equivariance, or student-teacher network models are trained with self-ensembling in \cite{Cui_2019,Yu_2019} and with transform consistency in \cite{Li_2020}.
FixMatch combines consistency regularization and pseudo-labeling to provide a simple yet powerful semi-supervised method for classification task \cite{sohn2020fixmatch}.
Few studies have explored the concept of FixMatch by adding discriminator to select pseudo-labels \cite{Mottaghi2020SemisupervisedSO} which adds extra complexity.
We introduce FixMatchSeg that provides transformation consistency required in semantic segmentation as in the case of \cite{Li_2020}, but without the complex teacher-student network.
Chaitanya et al. \cite{chaitanya2020contrastive} proposed a contrastive learning method to learn global and local features suitable for medical image segmentation with a small unlabeled dataset.
Contrastive self-supervised learning is complementary to semi-supervised learning, and hence can be used to pretrain before using semi-supervised approach, such as in \cite{pmlr-v119-chen20j} which combines SimCLR with FixMatch and named SelfMatch\cite{kim2021selfmatch} for classification task.

\section{FixMatchSeg}
{\label{sec:FixMatchSeg}}
Using the notations of FixMatch~\cite{sohn2020fixmatch} for consistency, let $X = \{(x_b,p_b):b \in (1,...,B)\}$ be a labeled data set of size $B$, where $x_b$ is an image and $p_b$ is the corresponding ground truth mask.
Let unlabeled batch be represented by a set $U = \{u_b : b \in (1,...,\mu B)\}$ where $\mu$ is a hyperparameter of the model determining the ratio of the unlabeled data to the labeled data.
Thus, we have $B$ labeled examples and $\mu B$ unlabeled examples in the training data.

Let $p_m(y|x)$ be the predicted class distribution image for the input image $x$. 
For semantic segmentation loss, we used the widely used soft dice loss $DL$ defined on \cite{shen2018influence} combined with boundary loss $BL$ explained in \cite{bokhovkin2019boundary}. In this study, both dice loss and boundary loss have equal contributions in the final loss calculation.

FixMatchSeg uses two types of image augmentations: strong denoted by $\mathcal{A}$ and weak denoted by $\alpha$. Unlike classification tasks, in semantic segmentation, the output target is not invariant under geometric transformations such as flips, affine or elastic distortions that change the shape or location of objects in the image.
Thus, for geometric transformation based augmentation, we apply the same transformation to both input image and mask label.
For weak augmentation $\alpha$, we chose random rotation and elastic distortions.
For strong augmentation $\mathcal{A}$, we modified the sharpness and contrast of the weakly augmented images and added Gaussian blur.
In order to have the same output target from the weakly augmented and strongly augmented images, instead of applying strong augmentation directly to the unlabeled image, we applied it to the weakly augmented version of that image. For the strong augmentation, we did not apply any kind of geometric or shape-changing transformation, which thus maintains the same geometrical shape of the objects in both weakly and strongly augmented images.

We have two kinds of losses: supervised loss $l_s$ and unsupervised loss $l_u$.
We compute supervised loss using soft dice loss $DL$ and boundary loss $BL$ against the labeled images, given by,
\[l_s = \frac{1}{B}\sum_{b=1}^B (DL\left(p_b,p_m\left(y|x_b\right)\right) + BL\left(p_b,p_m\left(y|x_b\right)\right))\]
where, $p_b$ is the ground truth label and $p_m$ is the predicted mask.

The key idea in FixMatch is to combine consistency regularization and pseudo-labeling into a single framework in unsupervised loss using weak and strong augmentations.
We followed the same idea in this work but adapt it to work for semantic segmentation tasks instead of classification. 
For each unlabeled input image $u_b$, we obtained an artificial label from its weakly augmented version $\alpha(u_b)$. For this, we first computed $p_m(y|\alpha(u_b))$, which is the model predicted class distribution map for the weakly augmented unlabeled image.
For $L$ class segmentation problem, each pixel of the image $p_m(y|\alpha(u_b))$ contains $L$ softmax output probabilities.
We compute pixel-wise max of this image to obtain $q_b$, and compute pixel-wise ${argmax}$ to obtain pseudo-label, $\hat{q_b} = \mathrm{argmax}[p_m(y|\alpha(u_b))]$.
Thus, $\hat{q_b}$ is the segmentation output predicted by the model for the weakly augmented unlabeled image.
In order to decide whether to use $\hat{q_b}$ as a pseudo-label or not, we first compute the confidence score as the mean of the pixel values of $q_b$, denoted by $\overline{q_b}$ which gives us the average maximum confidence of the model in predicting different classes over the whole image.
Note that while $\hat{q_b}$ is an integer image, $\overline{q_b}$ is a scalar number. If the average $\overline{q_b}$ is higher than the confidence threshold $\tau$, it is considered as a pseudo-label, i.e. ground truth for the strongly augmented unlabeled image $\mathcal{A}(\alpha(u_b))$.

Thus, we have unsupervised loss defined as follows:
\[l_u = \frac{1}{\mu B}\sum_{b=1}^{\mu B} \mathds{1}\left(\overline{q_b} \ge \tau \right)\ (DL\left(\hat{q_b}, p_m \left(y|\mathcal{A}\left(\alpha(u_b)\right)\right)\right) + BL\left(\hat{q_b}, p_m \left(y|\mathcal{A}\left(\alpha(u_b)\right)\right)\right)) \]

Then, the total loss for FixMatchSeg, $l$ is given by $l = l_s + \lambda_u l_u$, where $\lambda_u$ is the weight for the unsupervised loss.

\section{Implementation Details}
{\label{sec:implementation}}
\paragraph{\textbf{Datasets:}}
We applied our proposed method in four different medical image datasets, having different imaging modalities, different anatomical regions and potentially different requirement of shape vs texture bias for the models.

\textbf{CAMUS dataset \cite{Leclerc_2019}:}
CAMUS (Cardiac Acquisitions for Multi-structure Ultrasound Segmentation) dataset contains echocardiograms from 450 patients in the training set and 50 patients in the test set.
We do not have access to the ground truth of the test set, and the segmentation output images must be submitted online to obtain the dice scores on this test set.
There are four 2-D echocardiographic images for each patient: two 2-chamber view images and two 4-chamber view images captured during end-systole and end-diastole.
In each image, there are four labels: 0 (background), 1(left ventricular cavity), 2(myocardium) and 3(left atrium cavity).
In the experiments, we split the training set of CAMUS dataset into train, validation, and our test sets with 1000, 400, and 400 images, respectively.

\textbf{ISIC 2017 dataset \cite{DBLP:journals/corr/abs-1710-05006}:}
ISIC 2017  dataset is a collection of dermoscopic images provided in 2017 during the International Skin Imaging Collaboration (ISIC) Challenge.
There are a total of  2000 images in training set, 150 images in validation set, and 600 images in test set.
We did not make any changes in the structure of the dataset for our experiments. There are two labels in the ground truth 0 (Background) and 1 (lesion). 

\textbf{REFUGE dataset \cite{ORLANDO2020101570}}
REFUGE dataset is provided by MICCAI challenge in 2020. It consists of a total of 1200 retinal fundus images, 400 each in training, validation, and test sets. There are three classes, background optical cup and optical disc, to be segmented. 

\textbf{SCR dataset \cite{VANGINNEKEN200619}:}
SCR database is a dataset created for chest X-ray segmentation by annotating the radiographs available in the JSRT database.
It has a total of 247 images.
This dataset aims to segment the heart, lungs, and clavicles in the X-ray image. 

\paragraph{\textbf{Network Architecture and Hyperparameters:}}
We used a popular U-Net \cite{Ronneberger_2015} architecture with EfficientNet-B4 \cite{tan2019efficientnet} model as an encoder, which is a strong baseline for supervised semantic segmentation task and is available in Segmentation Models library \footnote{\url{https://github.com/qubvel/segmentation_models.pytorch}}
The model was implemented in the Python environment with Pytorch.
For optimization, Adam optimizer with the learning rate 0.001 was used.
For each experiment, the criteria for early stopping was that validation loss should not decrease for 9 consecutive epochs.
All the images were resized to 320x320.
The batch size is not constant for all experiments, it is dependent on the labeled to unlabeled example ratio $\mu$ was chosen.
For example, if number of labeled example is 10 and unlabeled is 100, then in this case we have a batch size of 11, where there will be one labeled image and 10 unlabeled images.
The confidence threshold $\tau$ was 0.90 for all experiments except those mentioned in Table 2.

\paragraph{\textbf{Data Augmentation:}}
We used imgaug library for image augmentation \footnote{\url{https://github.com/aleju/imgaug}}.
For weak augmentation, the images were rotated randomly in the range of $[-20, 20]$ degrees followed by an elastic distortion.
For strong augmentation, instead of original images, the weakly augmented images were taken as input and the sharpness and contrast of the images were modified followed by a Gaussian blurring.

\section{Results}
\label{sec:Results}
Table~\ref{tab:camus-results} compares supervised baseline with FixMatchSeg for CAMUS dataset when using 4, 8, 16, 32, 64, and 100 labeled images.
The results show that FixMatchSeg performs better than the baseline in most cases when few labeled images are used.
We see that usually, the dice score of FixMatchSeg prediction improves when the number of unlabeled images are increased with increasing $\mu$.
For 100 labeled images, we could not use ($\mu >  10$) as we have only 1000 images in our training set.
In our experiment, we have used augmented version of labeled data as a part of unlabeled data by removing its label.
Therefore, we were able to get results for $\mu = 10$.
We also compare the results when using transfer learning.
As shown in Table~\ref{tab:camus-results}, even with pre-trained network, FixMatchSeg provides very competitive results.
Table~\ref{tab:camus-threshold} shows the impact of changing threshold for selecting pseudo labels for different number of labeled images.
We see that at higher threshold levels from 80 to 100, the threshold value does not have a significant effect on the dice scores.

\begin{table}
\caption{{DSCs for baseline and FixMatchSeg for different number of labeled images for CAMUS dataset. For
FixMatchSeg,~\(\lambda_u\)=1,~\(\tau\)= 0.90.
{\label{tab:camus-results}}%
}}
\centering

\begin{tabular}{|c|c|c|c|c|c|c|}
\hline
model/labeled examples & 4 & 8 & 16 & 32 & 64 & 100 \\
\hline
Supervised baseline & 0.721 & 0.795 & 0.832 & 0.875 & 0.896 & \textbf{0.918} \\
FixMatchSeg ($\mu = 10$) & 0.736 &	0.781 &	0.\textbf{843} &	0.852 & 0.904 & 0.911 \\
FixMatchSeg ($\mu = 15$) & 0.729 & \textbf{0.807} & 0.842 & \textbf{0.888} & \textbf{0.906} & - \\
FixMatchSeg ($\mu = 20$) & 0.729 & 0.801 & - & - & - & - \\
FixMatchSeg ($\mu = 25$) & \textbf{0.756} & - & - & - & - & - \\
\hline
Supervised (pre-trained) & 0.87 & 0.91 & 0.92 & 0.93 & 0.94 & 0.94\\
FixMatchSeg (pre-trained) ($\mu = 9$) & \textbf{0.88} & 0.91 & 0.91 & 0.92 & 0.93 & \textbf{0.95}\\
\hline
\end{tabular}
\end{table}

In Table~\ref{tab:camus-testset}\, we compare the results of other state-of-the-art supervised network on CAMUS dataset, using an independent test set whose annotations are not available to us. 
To compute the results in this test set, we submitted the predictions from our model to the website \footnote{ \url{http://camus.creatis.insa-lyon.fr/challenge/#phase/5ca211272691fe0a9dac46d6}}, and we were provided with the results.

We could not compare several models with different parameters with this test set because we can submit the test result only four times for testing on the CAMUS website.
Among the three supervised models used by Sarah et al. \cite{Leclerc_2019}, the performance of U-net 2 is similar to the performance of our semi-supervised model trained with 100 labeled data.

\begin{table}
\caption{{Effect of choice of confidence threshold (\(\tau\)) and number of labeled images in FixMatchSeg for CAMUS dataset, where \(\mu\)=9.
{\label{tab:camus-threshold}}%
}}
\centering
\begin{tabular}{|c|c|c|c|c|c|}
\hline
labeled examples/\(\tau\) & 80 & 85 & 90 & 95 & 100 \\
\hline
4 & 0.8769 & 0.8768 & 0.8766 & 0.8772 & 0.8771
 \\
8 & 0.9114 & 0.9106 & 0.9118 & 0.9113 & 0.9122
 \\
16 & 0.9231	& 0.9227	& 0.9219	& 0.9205	& 0.9218
 \\
32 &	0.9214 &	0.9263 &	0.9261 &	0.9233 &	0.9246
 \\
64 &	0.9390 &	0.9371 &	0.9358 &	0.9399 &	0.9411
 \\
100	 & \textbf{0.9411} &	\textbf{0.9430} &	\textbf{0.9440} &	\textbf{0.9431} &	\textbf{0.9447}
\\
\hline
\end{tabular}
\end{table}

\begin{table}
\caption{{Results on the CAMUS test set submitted and obtained via an online portal and comparison with other's model performance in terms of dice score.Here, ED and ES refers to End-Diastolic and End-Systolic event of heart. The dice score for each individual structure is shown below. Segmented 3 cardiac structures are endo:endocardium, epi: epicardium and la: left atrium cavity. 
{\label{tab:camus-testset}}%
}}
\centering
\begin{tabular}[c]{|c|c|c|c|c|c|c|c|c|c}
\hline
& \multicolumn{6}{|c|} {Supervised training, 1624 labels} & \multicolumn{2}{|c|} {Semi-supervised, 100 labels}\\
\hline
 & \multicolumn{2}{|c|} {U-net-1 \cite{Leclerc_2019}} &  \multicolumn{2}{|c|} {ACNNs \cite{Leclerc_2019}} & \multicolumn{2}{|c|} {U-net-2 \cite{Leclerc_2019}} &  \multicolumn{2}{|c|} {FixMatchSeg (proposed)}  \\
\hline 
 & ED & ES & ED & ES & ED & ES & ED & ES \\
\hline 
endo & 0.94 & 0.91 & 0.94 & 0.91 & 0.92 & 0.9 & 0.92 & 0.88 \\
\hline
epi & 0.96 & 0.95 & 0.95 & 0.95 & 0.93 & 0.92 & 0.94 & 0.93 \\
\hline
la & 0.89 & 0.92 & 0.88 & 0.91 & 0.85 & 0.89 & 0.85 & 0.89 \\
\hline
\end{tabular}

\end{table}\par\null

\begin{table}
\centering
\caption{{DSCs for pretrained and randomly initialized baseline and FixMatchSeg forISIC 2017, SCR and REFUGE datasets. $Sup_{rand}$  and $Sup_{pre}$ refer to baselines with randomly initialised weights and pretrained weights respectively. Similarly, $SemiSup_{rand}$  and $SemiSup_{pre}$ refer to FixMatchSeg with randomly initialised weights and pretrained weights respectively.
FixMatchSeg,~\(\lambda_u\)=1,~\(\tau\)= 0.90.
{\label{tab:other-datasets-results}}%
}}
\begin{tabular}{|c|c|c|c|c|}

\hline
Datasets & $Sup_{rand}$ & $SemiSup_{rand}$ & $Sup_{pre}$ & $SemiSup_{pre}$\\
\hline
ISIC 2017 (100 labeled, 1000 unlabeled) & 0.72 & 0.66 & 0.74 & \textbf{0.76} \\
\hline
SCR (100 labeled, 100 unlabeled) & 0.84 & 0.85 &	0.86 & \textbf{0.87} \\
\hline
REFUGE (100 labeled, 400 unlabeled) & 0.50 & 0.79 & 0.58 & \textbf{0.85}\\
\hline
\end{tabular}
\end{table}

Table~\ref{tab:other-datasets-results} show results of FixMatchSeg in REFUGE, ISIC 2017 and SCR datasets using 100 labeled examples in each case. 
Compared to a supervised baseline, FixMatchSeg performs better in all the datasets.
The results also show that pre-training helps not just the supervised baseline but also semi-supervised model FixMatchSeg.
The improvement of FixMatchSeg over supervised baseline is very minor in ISIC 2017 and SCR, but quite significant in REFUGE dataset.

\begin{figure}[hbt!]
    \centering
    \includegraphics[width=0.5 \textwidth]{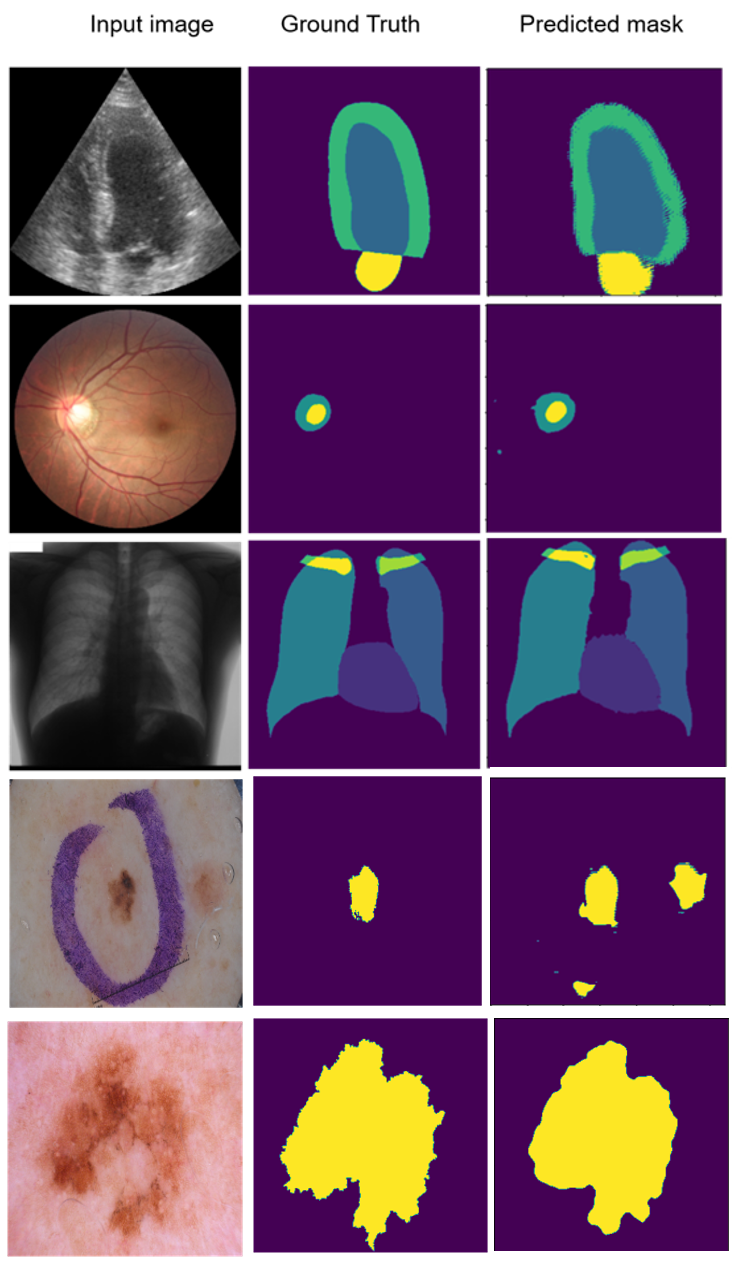}
    \caption{\textbf{Images with their ground truth and the segmented masks predicted by FixMatchSeg on CAMUS, REFUGE, SCR and ISIC 2017. The last row shows the image where the false segmentation mask is predicted because of a similar shape of both patches in the image.  }}
    \label{fig:results}
\end{figure}

\section{Discussion and Conclusion}
\label{sec:Discussion}

The evaluation of FixMatchSeg in four different datasets shows some promising results with few labeled data.
Even with the pre-trained network model with transfer learning, FixMatchSeg looks to be advantageous when we have few labels in the target domain.
However, the results and improvement over baseline depends on the dataset used and the ratio of labeled and unlabeled images.
In the same dataset, for example in CAMUS, we see that FixMatchSeg results improve as we increase the unlabled data.
However, it remains to be seen how and when this trend saturates as we keep on increasing the unlabled data.

When we compare ISIC 2017 and REFUGE datasets in Table~\ref{tab:other-datasets-results}, we see that although ISIC 2017 has greater proportion of unlabled images compared to REFUGE, FixMatchSeg improves supervised baseline by a much bigger margin in REFUGE.
This shows that only the number of unlabled image proportion is not a factor when we compare performance across datasets.
What factors impact the varied improvement in FixMatchSeg for different datasets?
We have used the same set of augmentation for all datasets.
Perhaps the interplay of augmentation, loss functions, pseudolabel threshold, and texture vs shape bias requirement for different segmentation task are complex that needs further exploration.
For example, the skin lesions are mostly irregular shaped but texture is curcial while for heart and lungs datasets, the regular shape plays an important role compared to texture (see Figure~\ref{fig:results}).
Moreover, there is still an open question on the impact of the diversity of unlabeled images that gets increased.
That is, what happens if we increase unlabeled data with only those images that are very similar to each other (a subdomain within a domain) vs images that are diverse in nature (covering the larger region of the data distribution of the domain)?
We will explore these questions in future work.

\textbf{{Acknowledgment}}

{This research was supported in part through computational resources provided by the Supercomputer Center Kathmandu University, which was established with equipment donated by  CERN. }

\bibliographystyle{splncs03}
\bibliography{references.bib}

\end{document}